\documentclass[letterpaper]{article} 
\usepackage{aaai25}  
\usepackage{times}  
\usepackage{helvet}  
\usepackage{courier}  
\usepackage[hyphens]{url}  
\usepackage{graphicx} 
\usepackage{natbib}  
\usepackage{caption} 
\usepackage{algorithm}
\usepackage{algorithmic}

\usepackage{amssymb}
\usepackage{multirow}
\usepackage{booktabs}

\usepackage{newfloat}
\usepackage{listings}
\DeclareCaptionStyle{ruled}{labelfont=normalfont,labelsep=colon,strut=off} 
\floatstyle{ruled}
\newfloat{listing}{tb}{lst}{}
\floatname{listing}{Listing}
\title{Addressing Bias in LLMs: \\Strategies and Application to Fair AI-based Recruitment}
\author{
    Alejandro Pe\~na,\textsuperscript{\rm 1}
    Julian Fierrez,\textsuperscript{\rm 1}
    Aythami Morales,\textsuperscript{\rm 1}
    Gonzalo Mancera,\textsuperscript{\rm 1}
    Miguel Lopez,\textsuperscript{\rm 1}
    Ruben Tolosana \textsuperscript{\rm 1}
}
\affiliations{
    \textsuperscript{\rm 1}Biometrics and Data Pattern Analytics Lab, Universidad Autonoma de Madrid, Spain\\

}

\begin{document}

\maketitle

\begin{abstract}
The use of language technologies in high-stake settings is increasing in recent years, mostly motivated by the success of Large Language Models (LLMs). However, despite the great performance of LLMs, they are are susceptible to ethical concerns, such as demographic biases, accountability, or privacy. This work seeks to analyze the capacity of Transformers-based systems to learn demographic biases present in the data, using a case study on AI-based automated recruitment. We propose a privacy-enhancing framework to reduce gender information from the learning pipeline as a way to mitigate biased behaviors in the final tools. Our experiments analyze the influence of data biases on systems built on two different LLMs, and how the proposed framework effectively prevents trained systems from reproducing the bias in the data.
\end{abstract}

%

\section{Introduction}

\label{sec:intro}

During the last decade, automatic decision-making systems have increased their presence in society, as AI-based technologies have continued to improve their performance in a vast diversity of tasks. A $2016$ report from the White House~\citep{white2016big} highlighted the key role that big data and automatic systems would play in high-stake domains such as employment, education or lending access. Since then, major developments have taken place in the field of AI. Specifically in the Natural Language Processing (NLP) domain, the Transformers model, based solely on self-attention mechanisms~\citep{vaswani2017attention}, supposed a paradigm shift with respect to previous recurrence-based models and paved the way for great models to come. Today, Large Language Models (LLMs) and foundational models such as novel GPT models~\citep{openai2023gpt4}, LLaMA~\citep{touvron2023llama2}, or DeepSeek, along with techniques that facilitate their fine-tuning to downstream tasks~\citep{hu2021lora}, enable unprecedented applications of this technology. 

Although NLP systems have proven to be an extremely useful tool for different applications\citep{pena2023leveraging}, concerns remain about how to properly use this technology to prevent harmful consequences. In this sense, in December $2023$ the European Parliament and the Council reached an agreement on the new AI Act, which establishes a risk-based categorization of AI systems, setting different requirements for each level. The final objective is to ensure the fundamental rights of EU citizens by creating an AI environment in which automatic systems deployed in Europe comply with the basic principles of a socially responsible AI~\citep{cheng2021socially}, such as privacy \citep{2022_Access_DP-CL_Ahmad}, or demographic fairness \citep{caliskan2017semantics,blodgett2020language,serna2022fair}. The recent emergence of foundational models and generative AI technologies has even highlighted the urgency of new regulations to address challenges related to these technologies~\citep{bommasani2023eu}, with the final objective of ensuring citizens' rights. In the US, a debate around the need to create new AI regulatory agencies is also taking place, with a focus on the concerns exposed.

In terms of the labor market and employment, the previously mentioned advances in AI at large have long contributed to this domain (e.g., AI-based personalized recommendations in employment platforms such as LinkedIn). Automatic decision systems have been of great use in different stages of the recruitment pipeline~\citep{bogen2018help}. Recent advances in prompt-based LLMs are encouraging the development of new tools\footnote{\url{https://www.herohunt.ai/blog/how-ais-like-gpt-are-changing-recruitment}} to help recruiters find the best suitable candidates for a job. As stated in the $2016$ White House report, the use of AI in employment presents some opportunities, for example, the prevention of human affinity biases during the screening or scoring of candidates. However, due to historical biases in the labor market~\citep{bertrand2004emily, bendick1997employment, altonji1999race}, a careful design of these systems with fairness perspectives is required~\citep{schumann2020we, sanchez2020does}. Some algorithms have exhibited demographic biases in this domain~\citep{amazon2018bias}, and therefore solutions to bias-free procedures are being proposed. These solutions range from addressing fairness in labor-related technologies, such as automatic ranking algorithms~\citep{yang2017measuring, zehlike2017fa}, to projects that focus on preventing discrimination in algorithmic hiring.\footnote{\url{https://www.findhr.eu/}}

In this work, we address the topic of demographic biases in Transformer-based technologies, with application to the case study of automated recruitment tools. With the aim of preventing data biases from affecting such systems, we present a privacy-enhancing learning framework that reduces gender information in the learning pipeline. In particular, we propose and evaluate two different methods to reduce demographic information in the system. One of these methods relies on deep model explainability, while the other is framed within the state of the art in bias mitigation in the image domain \cite{2020_ICPR_InsideBias_Serna,2022_SafeAI_IFBiD_Serna,nsigma}. We analyze the bias learned by a recruitment tool based on two different Transformers models, i.e. BERT~\citep{devlin2019bert} and RoBERTa~\citep{liu2019roberta}, and adapt the proposed methods to reduce gender information in the learning process. 

\section{Fairness Research in NLP}
\label{sec:related}

he subject of fairness and algorithmic discrimination in NLP has attracted great interest in recent years, following the general trend to achieve responsible systems in the AI community~\cite{cheng2021socially}.However, as some authors have pointed out~\citep{blodgett2020language}, most of the work that analyzes bias in NLP fails to properly define which behavior of the system is harmful, what is meant by a biased system \cite{serna2020aaai}, or the source of these biases. Several works have addressed gender bias in language technologies, due to the societal implications and prevalence of this kind of discrimination. For example, a great deal of attention has been paid to gender bias in language representations such as word embeddings~\citep{bolukbasi2016man,chen2021gender,garg2018word}. \citet{Arteaga2019Bias} analyzed different semantic representations and their relationship with the gender bias exhibited by occupation classifiers trained in Common Crawl biographies. They found strong gender biases that reflect historical biases in the labor domain using all three representations. Both~\citet{caliskan2017semantics} and~\citet{bolukbasi2016man} proposed bias metrics to analyze word embeddings, which they applied to the same pre-trained embeddings simultaneously, i.e. GloVe  embeddings, arriving to similar conclusions. In addition to gender bias,~\citet{caliskan2017semantics} measured human-like biases in word embeddings, finding that these captured commonly human prejudices well documented in psychology.  As showed by~\citep{garg2018word}, word embeddings trained on $100$ years of text data quantify historical demographic stereotypes, such as occupation rates. Social biases were also found in Transformers-based encoder representations~\citep{may2019measuring}, including intersectional biases such as the black woman stereotype.

Going beyond stereotypical associations in semantic representations, different works have analyzed demographic biases in various NLP tasks, including natural language inference~\citep{rudinger2017social}, language generation~\citep{sheng2019woman}, sentiment analysis~\citep{kiritchenko2018examining}, hate speech detection~\citep{sap2019risk, dixon2018measuring, park2018reducing}, machine translation~\citep{stanovsky2019evaluating, prates2020assessing}, or co-reference resolution~\citep{zhao2018bias, cao2021toward}, among others. As we aforementioned, different authors have pointed out the social perspectives and stereotypes depicted in text datasets as the source of these biases. A good example of this relationship can be found in the work analyzing social biases in toxicity and the detection of hate speech, where identity terms associated with certain demographic groups that are the target of hateful speech itself, such as women~\citep{park2018reducing} or queer people~\citep{dixon2018measuring}, are overrepresented in toxic samples. An unbalanced representation of identity terms is not the only way to introduce bias in a dataset, as noted in~\citep{sap2019risk}, where racial disparities were found to be a consequence of annotation biases due to the use of the African-American English dialect. It should be noted that some work has proposed methods to mitigate the biases learned by NLP models, such as removing sensitive indicators from input text~\citep{Arteaga2019Bias}, reducing the correlation between model output and embeddings of words that represent sensitive words~\citep{romanov2019whats}, data augmentation techniques~\citep{qian2022perturbation}, or even relying on adversarial prompting and safe response generation~\citep{ge2023mart}.

As argued by Blodgett \textit{et al.}, one recurrent problem in NLP works dealing with fairness is the vague description on how they conceptualize bias, and how the system behaviors can be harmful and to whom~\cite{blodgett2020language}. For instance, Caliskan \textit{et al.}, argued that the codification of a prejudice (i.e., what can be considered as a bias) is not harmful \textit{per se}, but it is the application context that determines whether that behavior would be harmful~\cite{caliskan2017semantics}. In general, we can categorize harmful system behavior into three different categories, namely: \textit{i) Representational harm}, which occurs when the system represents certain groups in a less favorable manner, or directly erases them ; \textit{ii) allocational harm}, when the system unfairly assigns resources differently across groups; and \textit{iii) quality of service harm}, which can occur when a system does not work the same for different groups. Although the varying performance of a sentiment analysis model depending on the demographics in the data samples could be a clear example of the latter~\cite{kiritchenko2018examining}, the prejudices encoded in word embedding representations as we aforementioned can suppose representational harm in situations where these stereotypical associations are reproduced. Our work here focuses on a case of allocational harm, where the demographic bias learned by the system could harm certain demographic groups by withholding the opportunities of those individuals to access a job. 

\section{Problem Formulation}
\label{sec:learning}

\begin{figure*}
\centering
    \includegraphics[width = \textwidth]{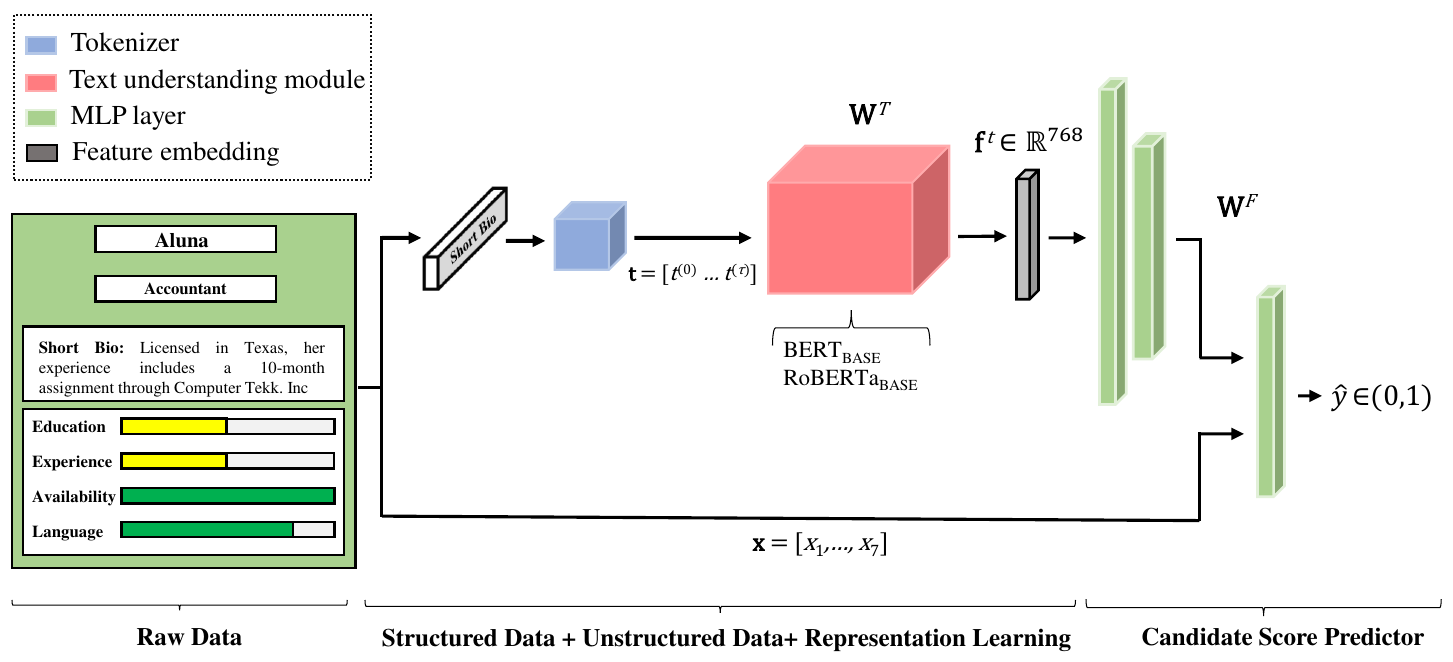}
    \caption{Learning architecture for the resume scoring tool $\mathbf{W}^S$. The system receives a multimodal input resume, composed by a candidate competencies vector $\mathbf{x}$ and a biography $\mathbf{t}$. A text understanding module $\mathbf{W}^T$ processes the bios to obtain a feature vector $\mathbf{f}^t$. Then, both $\mathbf{f}^t$ and $\mathbf{x}$ are fed to a fusion module $\mathbf{W}^F$ to obtain the final score $\hat{y}$.}
    \label{fig:learning-framework}
\end{figure*}

The aim is to train a fair scoring tool $\mathbf{W}^S$, based on structured and non-structured textual information. Concretely, suppose a multimodal dataset of resumes $\mathcal{D}$, where the $i$-th resume is composed of a vector of attributes or candidate competencies $\mathbf{x}_i$, and a biography, represented by a sequence of $\tau$ tokens $\mathbf{t}_i = [t^{(1)}_i,..., t^{(\tau)}_i]$ after tokenization. The resumes have a corresponding score $y\in[0,1]$, assigned by human annotators. Furthermore, we have access to the gender of the applicant $i$, which is a binary attribute $z_i\in\{0,1\}$ ($0$ for men, $1$ for women).\footnote{Although we adopt a gender definition that aligns in some points (e.g., gender as a binary attribute) with traditional definitions of such~\cite{keyes2018misgendering}, our intention is to improve the research on fairness in AI, for which we rely in existing data with such limitations, while considering the research conducted here to be extrapolable to other definitions. As some authors have pointed out, a crucial step in preventing social injustice and demographic bias in AI is to properly define these terms, and consider the perspectives of individuals belonging to these groups~\citep{leavy2021ethical}, for which a shift to new, inclusive definitions is needed~\cite{cao2021toward}.} In this context, the scoring tool $\mathbf{W}^S$ could be built on the general learning architecture of Fig.~\ref{fig:learning-framework}. This system is trained in $\mathcal{D}$ to correctly predict a new score $\hat{y}_i$ that approximates $y_i$ based on an input resume $\{\mathbf{x}_i,\mathbf{t}_i\}$. Note that the scoring tool is composed of: \textit{i)} a text understanding module $\mathbf{W}^T$, which takes a tokenized biography as input and outputs a text embedding $\mathbf{f}^t \in\mathbb{R}^N$; and \textit{ii)} a fusion module $\mathbf{W}^F$, which further processes $\mathbf{f}^t$, and merges it with candidate competencies ($\mathbf{x}$) to predict a score $\hat{y}$.

Suppose now that the scores are not a completely objective assessment of the candidate but a certain degree of subjectivity was included during their generation. Without loss of generality, suppose that human annotators assigned lower scores to women compared to men. Subjectivity during annotation leads to a new set of gender-biased scores $y^G$. In this context, a data-driven learning framework could exploit the demographic information in the biographies and use it to learn bias in scores $y^G$, leading to a scoring tool that unintentionally discriminates by gender. Using these scores to select potential candidates for a job, for example, by providing a short list of the best suitable applicants, would result in a case of allocational harm~\citep{blodgett2020language}, as the scores $\hat{y}$ calculated by the tool would unfairly assign better opportunities to one demographic group using information about the sensitive attribute $z$ (i.e., under the same conditions, the score for men would be higher than the score for women). In order to prevent this behavior from the recruitment tool to unfairly discriminate again a demographic groups, we explore here the application of a privacy-enhancing framework as a bias mitigation strategy.

\section{Bias Reduction: Proposed Methods}
\label{sec:methods}

\begin{figure*}
\centering
    \includegraphics[width = \textwidth]{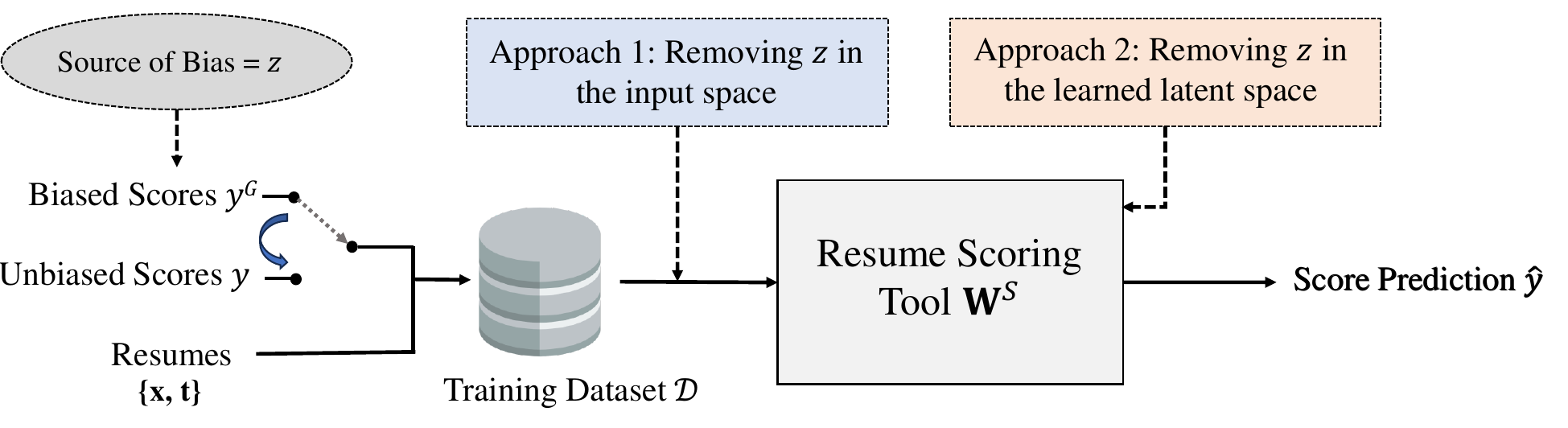}
    \caption{Proposed privacy-enhancing learning framework for bias mitigation. We aim to mitigate the effect of demographic data biases by removing demographic information in the input space or in latent space of an AI-based resume scoring tool.}
    \label{fig:block_diagram_approaches}
\end{figure*}

We propose to remove gender information from the previously introduced learning framework with two different approaches (see Figure \ref{fig:block_diagram_approaches}). Each method focuses on a different stage of the learning process. The idea is that, by removing gender information, the network will not be able to establish a relationship between the sensitive attribute $z$ (gender in this case) and the bias in the scores, therefore preventing it from reproducing the bias.

\subsection{Via Model Explainability}

This approach performs an iterative learning process in which potential biased tokens are detected and removed from the input sentences in each iteration, followed by a network retraining. This could be achieved by applying the Integrated Gradients technique to the model~\citep{sundararajan2017axiomatic}. Integrated Gradients (IG) is a model explainability technique that allows us to determine the importance of input features in the model's predictions, without requiring us to modify the model. This method begins with a baseline input $\mathbf{f}^{'}$, which could be a zero vector, and builds a sequence of data samples as the interpolation between $\mathbf{f}^{'}$ and the input at hand $\mathbf{f}_i$. For each data sample, the gradients are computed to measure how changes in the features affect the model's prediction. Gradients are accumulated, resulting in an attribution mask that resembles the relevance of the characteristic.

This technique measures token relevance in a validation set and detect relevant words in prediction that do not appear to correlate with the target task or that explicitly correlate with the sensitive attribute. Our hypothesis is that, in the presence of data biases that affect the predictions of the model, some tokens will act as proxies for the sensitive attribute $z$, thus appearing as relevant tokens for the predictions. This approach improves the process in a data-driven way with respect to~\citet{Arteaga2019Bias}, where, after removing gendered pronouns and names from biographies, the authors concluded that some tokens acted as proxies for the gender attribute. Thus, measuring token relevance in a biased prediction context could be useful for detecting words that act as such proxies but have little impact on the main task. 

\subsection{Via Adversarial Learning}

Our second approach is based on domain-adversarial learning for image analysis applications. We propose adapting the method of~\citep{kim2019Learning}, namely learning-not-to-learn (LNTL), to remove gender information from hidden representations of the scoring tool $\mathbf{W}^S$. Originally proposed to prevent a network from learning a data bias unrelated to the main task during training, the method can be applied as a privacy enhancing technique to remove sensitive information from latent representations \cite{pena2021emotion,Morales2021SensitiveNets}. This method proposes to reduce the mutual information between the hidden representation of a network and the data bias through the following loss function that we adapt to our case:

\begin{equation}
\label{eq:lntl_adapted}
    \mathcal{L}_{\textrm{RMSE}}(\{\mathbf{x},\mathbf{t}\},y|\mathbf{W}^{S}) + \lambda \mathcal{I}(z;f(\{\mathbf{x},\mathbf{t}\}|\mathbf{W}^{S}))
\end{equation}

\noindent where $\mathcal{L}_{\textrm{RMSE}}$ is the Root Mean Squared Error Loss (i.e., we consider the score prediction task as a regression problem), $\mathcal{I}$ is mutual information, and $f(\{\mathbf{x},\mathbf{t}\}|\mathbf{W}^{S})$ is a latent hidden representation of $\mathbf{W}^S$, from which we want to remove the information. The mutual information in Eq.~\ref{eq:lntl_adapted} is reduced by optimizing the negative conditional entropy $-\mathcal{H}(z|f(\{\mathbf{x},\mathbf{t}\})$, for which the \textit{a posteriori} distribution is required. The authors of LNTL proposed to attach an auxiliary classifier to the network in order to predict the bias (i.e. $z$ here) from the latent representations and use its output to approximate such distribution. In the context of the learning architecture of Fig.~\ref{fig:learning-framework}, the classifier would be attached to the output of an intermediate layer of $\mathbf{W}^F$. To train that intermediate classifier, here a min-max adversarial learning framework~\citep{goodfellow2014gan} is used, where we are simultaneously learning the main task (i.e. score prediction), while using an auxiliary prediction network to reduce gender information from latent representations of $\mathbf{W}^F$. Consequently, the characteristics extracted by $\mathbf{W}^S$ make it increasingly difficult for the auxiliary classifier to predict $z$ as training progresses, that is, our original objective. 

\section{Materials and Methods}

\subsection{Dataset: FairCVdb}
\label{sec:material}

FairCVdb is a database for multimodal fairness research in AI~\citep{pena2020cvprw}, with which various fairness testbeds have recently been defined \citep{pena2020icmi,pena2023human}. The database comprises $24$,$000$ synthetic profiles, which were generated by using the identities contained in the DiveFace Face Recognition database~\citep{Morales2021SensitiveNets} and the biographies of~\citep{Arteaga2019Bias}. Each profile includes the following information, usually present in a resume:

\begin{itemize}
    \item \textbf{Non-structured Textual Information}: Each sample comprises two short biographies (a raw-text bio and a gender-agnostic bio), an occupation, and a name. FairCV spans$10$ different occupations from $4$ different labor sectors.
    \item \textbf{Structured Information}: FairCV includes $7$ features from the $5$ information blocks, including education, availability, previous experience, recommendation, and language proficiency. These features are known as candidate competencies. In addition to the competencies, gender is also available.
\end{itemize}

 Each resume was scored using a weighted linear combination of the candidate's competencies and the labor sector (weights defined by experts on human resources). Although these scores were generated without any consideration of demographic information, thus becoming blind scores, the authors generated two additional sets of scores that included demographic biases, i.e., gender- and ethnicity-biased scores. For the scope of this work, we are only interested in using blind and gender-biased scores. Furthermore, we consider the blind scores as the ground truth of the resumes.

\subsection{Language Models}
\label{sec:models}

We will employ two different language understanding models to process the biographies of FairCVdb:

\begin{itemize}
    \item \textbf{BERT}~\cite{devlin2019bert}: The BERT model was the first transformed-based language model~\cite{vaswani2017attention} to apply bidirectional training to learn language representations. Thus, BERT is able to attend to both the left and right context in the sentence (i.e., bidirectional self-attention), and process the whole input at once. In order to achieve this, BERT employs a Masked-Language Modeling (MLM) pre-training objective, where arbitrary tokens are masked, and the network has to predict them based on the surrounding context. Along with the MLM objective, they include a Next Sentence Prediction (NSP) loss to improve the model’s capacity to understand the relationship between two different sentence. Two versions of BERT are available, the base version ($12$ layers, with $12$ attention heads, and a hidden size of $768$ features) and the large version ($24$ layers, with $16$ attention heads, and hidden size of $1024$ features).
    \item \textbf{RoBERTa}: The RoBERTa model is a variant of the BERT model. ~\cite{liu2019roberta} conducted a study on the BERT pre-training strategy, in which they found that it was indeed undertrained. They used the same architecture, and pre-trained the model in a larger text corpora, carefully selecting the hyperparameters (e.g., a larger batch size). Another key finding of such work was the performance improvement in the downstream task by removing the NSP loss from the pre-training objective. Consequently, the model was pre-trained using only MLM loss. Whilst in BERT's implementation for the MLM loss tokens were randomly masked once during preprocessing, RoBERTa's training strategy allows to randomly mask tokens in each iteration, hence obtaining a much larger corpus of potential masking combinations. Similarly to BERT, RoBERTa is available in the same base and large versions.
\end{itemize}

\subsection{Fairness Criteria}

In this work, we train a fair resume scoring tool in the presence of gender biases (originated due to annotation bias in the dataset). When implemented, the resume scoring system would assign a predicted score $\hat{y}$ to applicant resumes to help human recruiters find the most suitable candidates for a job. One way to do this is by computing a shortlist of the best suitable candidates, sorted by $\hat{y}$. Note that while the scoring tool itself is a regression model (i.e., it predicts a continuous score $\hat{y}\in[0, 1]$), the recruitment process can be modeled as a binary classification problem, in which the predicted score $\hat{y}$ is the confidence used to determine whether the candidate appears on such a shortlist. In order to assess the (un)fairness of the whole process resulting from the predictions $\hat{y}$, we use two fairness metrics commonly applied in traditional Machine Learning~\citep{hardt2016equality}:

\begin{itemize}
    \item \textbf{Statistical Parity}. This criterion requires the probability that a decision is independent of membership in sensitive groups. In our case, this leads to an equal probability of appearing in the shortlist across genders, which can be approximated by the gender proportions within it. Failing to achieve this criterion suppose an \textit{evidence} of allocational harm, but it does not conclude so in general since it does not consider features relevant to the decision.

    \begin{equation}
    \label{eq:ch_fairness_survey_sp}
        P(\mathrm{\hat{y}} = 1 | z = 0) = P(\mathrm{\hat{y}} = 1 | z = 1)  
    \end{equation}
    
    \item \textbf{Equality of Opportunity}. This criterion is commonly used in classification tasks in which the positive outcome is associated with an advantaged decision. The criterion ensures that the automatic process performs equally well across sensitive groups. To do so, it requires independence of the sensitive attribute subject to a positive ground truth (i.e. similar recall), measuring only whether individuals that deserve the positive outcome are treated equally. 

    \begin{equation}
    \label{eq:ch_fairness_survey_eop}
        P(\mathrm{\hat{y}} = 1 | z = 0, \mathrm{y} = 1) = P(\mathrm{\hat{y}} = 1 | z = 1, \mathrm{y} = 1), 
    \end{equation}
\end{itemize}
    
In the US employment law domain, a commonly applied discrimination measure is the $4/5$ rule, a heuristic rule that states that if the hiring rate of a demographic group is less than $0.8$ of the group with the highest hiring rate, there is evidence of disparate impact~\cite{us1979questions}. Consequently, the rule is a direct measure of statistical parity, as it measures whether a statistical parity ratio (i.e., the ratio obtained when comparing the proportions of two demographic groups) exceeds the $0.8$ threshold. This rule operates as a ``rule of thumb'', since failing to achieve it is not enough to demonstrate discrimination or unfairness~\cite{watkins2022four}, but rather to denote a \textit{prima facia evidence} that discrimination may be occurring. Nevertheless, we can use the demographic ratio as a fairness measure in the scoring tools trained with FairCV data, as the dataset groundtruth (i.e., the blind scores $y$) is completely balanced in terms of gender, with equal representation of each group. This should prevent our assessment from failing to consider gender-related contextual variables that affect the gender proportions in the predictions beyond the bias in $y^G$, thus preventing a portability trap in the application of this measure. Furthermore, we will use the recall computed for different groups to include a fairness measure that also considers the utility of the system.

\section{Experimental Setup and Results}
\label{sec:experiments}

We train the scoring tool using the architecture of Fig.~\ref{fig:learning-framework} using FairCVdb~\cite{pena2023human}. We explore two Transformers (Sect.~\ref{sec:models}) for the text understanding module $\mathbf{W}^T$, which are frozen during training. The feature vector $\mathbf{f}^t\in\mathbb{R}^{768}$ is obtained as the average of all the output embeddings. Taking into account $\mathbf{W}^F$, we implement it as a naive MLP with three layers. The first two layers, with $300$ and $20$ units, further process $\mathbf{f}^t$. Then we concatenate the output of these layers with the candidate competencies $\mathbf{x}$, and compute the final score $\hat{y}$ with an output layer. We use a sigmoid activation in all layers and a dropout rate of $0.3$. The system is trained for $10$ epochs using RMSE loss, batch size of $32$, AdamW optimizer ($\beta_1 = 0.9, \beta_2 = 0.999, \epsilon = 1e-8$), and $lr = 1e-3$. We use the train/validation splits from FairCVtest~\citep{pena2020cvprw,pena2020icmi,pena2023human}.

\begin{table*}[htb]
    \caption{Gender proportions in the top-$500$ candidates of the validation set of FairCVtest~\citep{pena2020icmi}. We obtained the shortlists of candidates by computing resume scores $\hat{y}$ with different configurations of the scoring tool.}
    \label{tab:rank}
    \centering
    \begin{tabular}{llccccccc}
    \toprule
    \multirow{2}{*}{\textbf{Model}} &\multirow{2}{*}{\textbf{Target}}&\multirow{2}{*}{$\mathbf{D_{KL}}$}&\multicolumn{3}{c}{\textbf{Proportion}}&\multicolumn{3}{c}{\textbf{Recall}}\\
    &&&\textbf{M}&\textbf{F}&\textbf{Ratio}&\textbf{M}&\textbf{F}&\textbf{Overall}\\
    \midrule
    \multirow{2}{*}{$\mathrm{BERT_{BASE}}$}&Unbiased&$0.0083$&$48.60\%$&$51.40\%$&$0.946$&$76.80\%$&$79.60\%$&$78.20\%$\\
    &Biased&$0.3102$&$69.40\%$&$30.60\%$&$0.441$&$87.60\%$&$54.80\%$&$71.20\%$\\
    \hline
    \multirow{2}{*}{$\mathrm{RoBERTa_{BASE}}$}&Unbiased&$0.0116$&$47.80\%$&$52.20\%$&$0.916$&$74.00\%$&$77.20\%$&$75.20\%$\\

    &Biased&$0.3387$&$67.00\%$&$33.00\%$&$0.493$&$83.60\%$&$56.00\%$&$69.80\%$\\
    \hline

    \end{tabular}
\end{table*}

\subsection{Gender Information Analysis}

In the first experimental part, we want to assess the capabilities of the system to detect information related to the sensitive attribute $z$ from the representations extracted by the transformer model. To do so, we propose a screening process based on $4$ different resume scoring systems. The idea is to use each system to assign a score to the resumes included in the validation set, then rank them according to the predicted scores $\hat{y}$, and extract gender proportions in the pool of suggested candidates. Each system in this experiment was trained using one of the two Transformer models and one of the two scores available on the resumes (i.e., unbiased $y$ and gender-biased scores $y^G$).

Tab.~\ref{tab:rank} presents the proportion of males and females in a shortlist with the top-$500$ candidates, according to the scores predicted by each system. Note that we also included the Kullback-Leibler divergence $D_{\textrm{KL}}$ between the gender distributions of the predicted scores as a quantitative measure of the gender bias of the scoring tools, as previously used by~\citet{pena2023human}. Observing the systems trained with the unbiased scores, we can observe a similar behavior using both language models. The proportions of males and females in the shortlist of candidates are nearly equal for the RoBERTa-based model, and even perfectly equal for the BERT-based model. In this context, systems learn to predict unbiased scores, as indicated by the low values of $D_{\textrm{KL}}$ across the distributions. The demographic ratio in the unbiased scenario is greater than $0.91$ for both models, further highlighting the similar presence between the groups in the shortlist. In terms of recall, the BERT-based scoring tool correctly includes $78.2\%$ of the candidates with the highest ground-truth score on its shortlist, a slightly higher value than the RoBERTa-based model (that is, $75.20\%$). If we observe the recall by gender, we see a gap shorter than $3$ points in both cases, with a higher recall for females.

However, we observe a clear difference when training models with gender-biased scores. Both scoring systems are clearly affected by the gender bias induced in the scores. This can be seen in how $D_{\textrm{KL}}$ increased over $0.3$ for the distributions of the scores predicted with both models. The proportion of males is $69.4\%$ for the BERT-based model and $67\%$ for the RoBERTa-based model, leaving the female rates at nearly $30\%$. In these conditions, the demographic ratios decrease to $0.441$ and $0.493$, respectively. Note that, according to the $4/5$ rule, a ratio lower than $0.8$ is considered evidence of discrimination. Recall that gender information is not included in the candidate competencies $\mathbf{x}$. Therefore, the systems are extracting gender information encoded within the deep representation $\mathbf{f}^t$ and exploiting it to fit the loss objective during training by learning the bias in the scores. In terms of recall, the higher presence of males in the shortlist led to an increase in the recall of this group, which raised about $10$ points for both systems. Naturally, the opposite occurred with women, for which a drop of more than $20$ points is observed, leading to a clear violation of the criterion of Equality of Opportunity. Interestingly, overall recall decreased significantly for both systems, highlighting the fact that the systems are not only discriminating against women, but recovering a lower percentage of suitable candidates.  

\subsection{Approach 1: Via Model Explainability}

As we have shown in the previous section, when training a scoring tool when gender bias is present in the objective function, both systems establish a correlation between the scores and the gender information encoded in the text representations extracted by the Transformer. Thus, the final system shows biased behavior towards this group when predicting new scores, reproducing the bias in the training set. To mitigate the effect of such bias, here we apply an iterative learning process based on the use of the Integrated Gradients~\cite{sundararajan2017axiomatic} technique to detect gendered words and remove them from the input before retraining the system (see Section~\ref{sec:methods}).

Starting from the biased systems that we trained in the previous section, we compute the attributions of the output with respect to the Transformers' embedding layer, and select the tokens based on the embedding relevance. We use a zero vector as reference for the candidate competencies and a sequence of [PAD] tokens as reference for the text input in each case. For each resume, we extract the $20$ most relevant tokens. We then extract the frequency of the tokens in this set of tokens by gender and labor sector, obtaining for each of the resulting $8$ groups a set of the $30$ most relevant tokens. To create these sets, we previously filtered stop words, adverbs, and verbs that provide little semantic information.

The first thing we notice when inspecting the sets is the presence of words related to the corresponding sector. This fact corroborates the usefulness of the method for detecting relevant words. We detect in the sets the gendered words manually removed by~\citet{Arteaga2019Bias}. We also find several words denoting potential social biases in the sets, such as \textit{children} in both the BERT and RoBERTa female set for the education sector, and the word \textit{family} in the BERT female set for the healthcare sector and in the RoBERTa female set for the jurisdiction sector. Although one could argue that \textit{children} is a word expected in the education sector, the absence of such a word in the corresponding male set is consistent with social stereotypes that place women in a caregiving position. The same could be argued for the case of \textit{family}. Interestingly, we also find the word \textit{husband} in BERT's female set for education. This word does not provide any meaning for the labor sector, and in fact can serve as a proxy for the gender of the bio (in a context where most women in the dataset are heterosexual). Other interesting words aligning with social gender stereotypes can be found in the education sector. Concretely, in the female set we encounter the words \textit{psychology} and \textit{health}, while in the male set we find \textit{technology, engineering,} and \textit{economics}. This distinction is in line with the social stereotypes about the prevalence of each group in different occupations. We note here that this kind of words may be those referred to in the conclusions of~\citet{Arteaga2019Bias}, since they correlate with gender by unequal representations instead of explicitly containing gender information.

We replace the previous tokens, as well as the bios names, with the BERT/RoBERTa mask token ([MASK]), and retrain the scoring tool $\mathbf{W}^S$ with the biased scores. In this case, both systems show an unbiased behavior in their predictions $\hat{y}$. Tab.~\ref{tab:rank_fair} presents the proportion of each gender in the top-$500$ candidate shortlist using the scoring tool trained with this approach. Despite being trained with the same biased target function as the baseline models, both genders have almost equal representation in these lists, exhibiting even higher demographic ratios than those obtained with the unbiased models (cf. Tab.~\ref{tab:rank}). By removing gendered words from the input bios, the gender information within the system was effectively reduced. Hence, trained systems cannot reproduce the bias, as they have no information about the cause (i.e., $z$-related membership) of the difference in scores. We can corroborate bias mitigation by looking at the predicted score distributions $\hat{y}$ by gender on the validation set. Although the $D_{\textrm{KL}}$ between the distributions is higher than $0.3$ for both baseline models, the value is reduced to $0.0267$ for the BERT-based tool and $0.0168$ for the RoBERTa-based one. 

Looking now at the utility of the systems, both systems present a higher overall recall than the biased scoring tools. The removal of gender information not only equalized the representation of both groups, but improved the retrieval of suitable candidates. By not distinguishing the source of the bias in the target function $y^G$, the scoring tools are forced to pay more attention to the rest of the features in order to fit the training objective. Hence, we have improved not only the fairness of the systems, but also their utility. Note here that the recall for males decreased compared to biased scenarios. The increase in recall in males when using biased tools was due to the greater presence of this group in the candidate shortlist, rather than to better performance. This can be corroborated by comparing the gender recall values obtained here with the values obtained with the unbiased tools, which are closer. However, the recall values obtained by the unbiased tools are slightly higher, indicating that even after mitigating the effect of gender bias, this method is still penalized for it.

\begin{table*}[t!]
    \caption{Gender proportions in the top-$500$ candidates of the validation set of FairCVtest. For both scoring tools, we extract results after removing gender information in the input space (Approach $1$) or in a latent space (Approach $2$).}
    \label{tab:rank_fair}
    \centering
    \begin{tabular}{llccccccc}
    \toprule
    \multirow{2}{*}{\textbf{Model}} &\multirow{2}{*}{\textbf{Method}}&\multirow{2}{*}{$\mathbf{D_{KL}}$}&\multicolumn{3}{c}{\textbf{Proportion}}&\multicolumn{3}{c}{\textbf{Recall}}\\
    &&&\textbf{M}&\textbf{F}&\textbf{Ratio}&\textbf{M}&\textbf{F}&\textbf{Overall}\\
    \midrule
    \multirow{2}{*}{$\mathrm{BERT_{BASE}}$}
    &Approach $1$&$0.0267$&$50.80\%$&$49.20\%$&$0.969$&$74.40\%$&$76.40\%$&$75.40\%$\\
    &Approach $2$&$0.0422$&$54.20\%$&$45.80\%$&$0.845$&$86.40\%$&$79.60\%$&$83.00\%$\\
    \hline
    \multirow{2}{*}{$\mathrm{RoBERTa_{BASE}}$}
    &Approach $1$&$0.0168$&$49.00\%$&$51.00\%$&$0.961$&$71.60\%$&$74.40\%$&$73.00\%$\\
    &Approach $2$&$0.0487$&$54.60\%$&$45.40\%$&$0.831$&$83.40\%$&$76.60\%$&$80.00\%$\\
    \hline

    \end{tabular}
    
\end{table*}

\subsection{Approach 2: Via Adversarial Learning}

In this section, we apply the method of~\citet{kim2019Learning} (LNTL) to remove gender information in latent representations of scoring tools $\mathbf{W}^S$ during training. Concretely, we will remove the gender information from the $300$ dimensional feature vector computed by the first layer of $\mathbf{W}^F$. To this end, we attach an auxiliary gender classifier to this layer. The gender classifier is made up of a hidden layer with $20$ units and an output layer with two output units, softmax activation. Similarly to $\mathbf{W}^F$, we set a dropout rate of $0.3$  and use a value of $\lambda = 0.1$ to balance the terms in Eq.~\ref{eq:lntl_adapted}.

\begin{figure*} [t!]
    \centering
    \includegraphics[width = \textwidth]{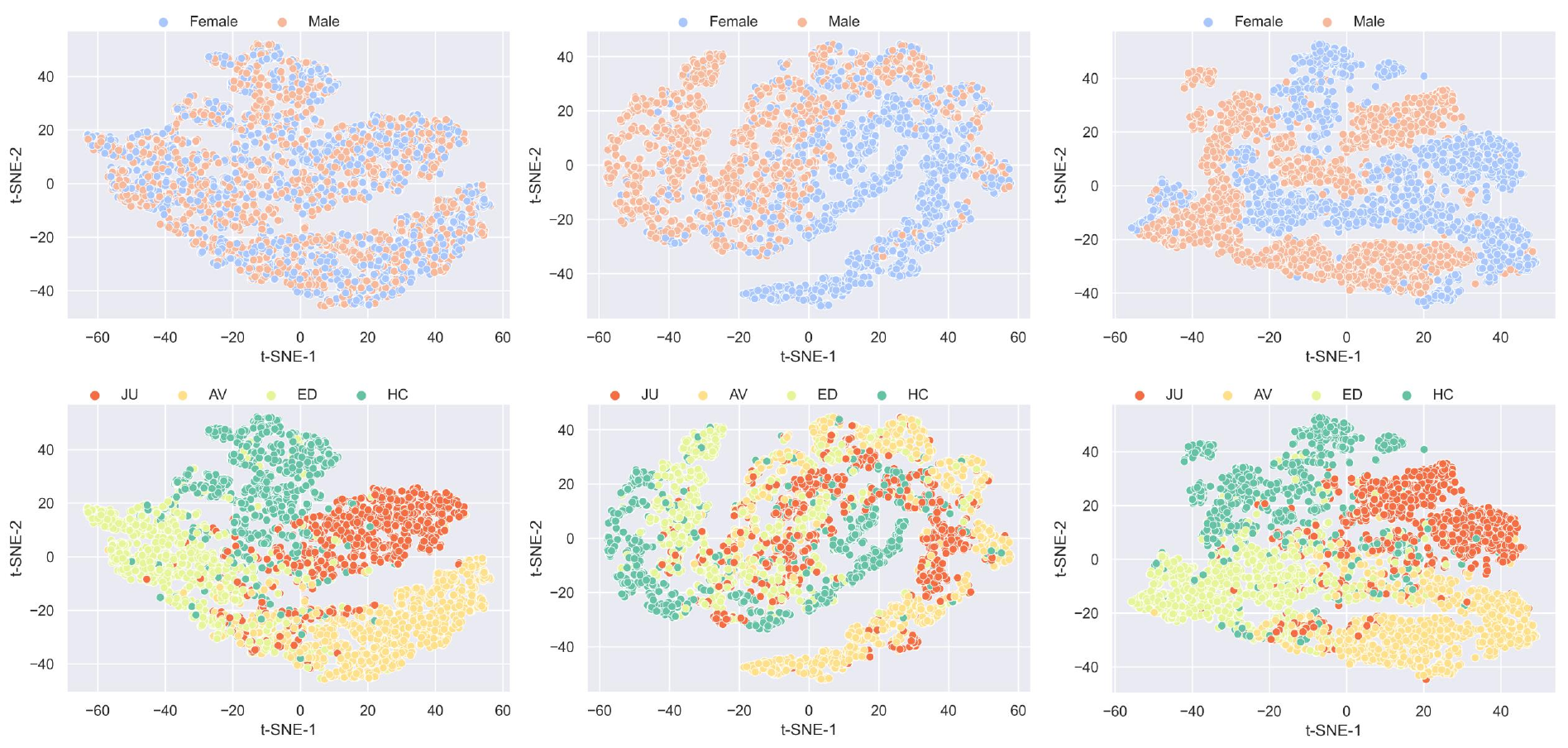}
    \caption{Visualization of the t-SNE projections for the $300$-dimensional feature embeddings computed with the BERT-based scoring tool in different setups: (Left) the baseline scoring tool trained with unbiased scores; (Center) the baseline scoring tool trained with biased scores; and (Right) the scoring tool trained with biased scores using the Approach $2$.}
    \label{fig:tsne}
\end{figure*}

The results of training the scoring tools with biased scores using this approach are presented in Tab.~\ref{tab:rank_fair}. Similarly to the results obtained with Approach $1$, the recruitment systems trained with the adaptation of LNTL significantly reduced the impact of bias. Looking first at the distributions of $\hat{y}$, both systems exhibit a $D_{\textrm{KL}}$ value under $0.05$, far from those obtained in the biased baseline scenario. However, the values obtained with the approach $2$ are far from the unbiased setup, and significantly higher than the ones of Approach $1$. This indicates that while this method has significantly reduced the effect of bias, it has not removed it as effectively as the previous approach. However, for both systems, the demographic ratios comply with the rule of $4/5$ and are significantly higher than at the baseline. Although the gender gap is greater here than when using Approach $1$ or the unbiased baseline, we note that the recall values obtained are the highest among all configurations. Surprisingly, both systems achieve recall values in the male group close to those obtained with the biased baselines, without the increased presence of this group in the shortlist. This fact indicates that the removal of latent gender information encouraged the tool to focus on the rest of the features, leading to a more accurate prediction of the score. 

To further analyze the impact of LNTL in the latent representations, we compute the t-SNE~\cite{van2008visualizing} projections of the $300$-dimensional latent vectors from the validation set of FairCV. This dimensionality reduction representation works on the basis that similar samples in the original spaces should be projected to nearby points in the low-dimensional space, while dissimilar samples should fall far apart. Thus, the representation is useful to understand relationships and data clusters in the original space, but do not have class semantic meaning, as it is an unsupervised method. We present the t-SNE projections computed for the BERT-based scoring tool in Figure~\ref{fig:tsne}, for which we provide annotations of gender (Upper row) and labor sector (Bottom row) after computing the representations. Figure~\ref{fig:tsne} (Center) presents the projection of the feature vectors extracted with the biased baseline, while Figure~\ref{fig:tsne} (Right) presents the proper when applying LNTL. We included as well in Figure~\ref{fig:tsne} (Left) the projections of the baseline tool trained with the unbiased scores $y$ for comparison purposes. Attending first to the unbiased tool, we appreciate that the representations ignore gender information, with the samples forming clusters based on the labor sector. Conversely, we observe that the labor clusters disappear in the embeddings extracted with the biased baseline tool, as the samples belonging to each labor sector appear dispersed (i.e., for instance, the Jurisdiction group), and even mixed. Instead, we can observe how most of the samples seems to group in two regions based on gender (Figure~\ref{fig:tsne} (Center-Up)). Despite not being explicitly trained to infer gender, due to this information being crucial to fit the biased target function, the latent representations capture gender information with even more relevance that labor-related information. When Approach $2$ is applied, an interesting projection is obtained. We notice that the labor-related information regains significance within the latent representation compared to the biased scenario, as the samples seem to group again by labor sector. However, if we pay attention to the projections annotated by gender, we can appreciate that, within each labor-cluster, the model is able to distinguish each gender group. Note that, by not forming general clusters as in the biased scenario, we cannot conclude that gender is the predominant information within the latent representation, but the tool clearly has some sense of it. This is consistent with the results presented in Table~\ref{tab:rank_fair}, in light of which we have already commented that gender information was not completely removed, as denoted by the gaps in the performance between groups.

\section{Conclusions}
\label{sec:conclusion}

We have addressed the topic of gender bias and fairness in LLMs with application to the labor market. We have used a synthetic profile dataset to evaluate the capacity of a Transformer-based scoring tool to exploit gender-related information in textual data. Our initial hypothesis was that, in the presence of a biased target function, the system would use the information in the text input to learn the bias. Hence, systems trained in this context would reproduce biased behavior toward a gender group.

We explored the behavior of two different LLMs, BERT and RoBERTa, in the context described above. In our experiments, we demonstrate how the systems based on both models effectively learn bias. We explore the application of privacy-enhancing methods to reduce gender information in the system as a way to prevent it from learning the bias. Concretely, our framework adapts two different methods, a deep explainability-based technique and a bias-aware learning method. Our experiments show how the application of both methods successfully reduced gender information in the pipeline, thus preventing biased behaviors despite training in the presence of data biases.

\section{Acknowledgments}
This study has been supported by the projects BBforTAI (PID2021-127641OB-I00 MICINN/FEDER) and Cátedra ENIA UAM-VERIDAS en IA Responsable (NextGenerationEU PRTR TSI-100927-2023-2). The work G. Mancera is supported by FPI-PRE2022-104499 MICINN/FEDER. The work has been conducted within the ELLIS Unit Madrid.

\bibliography{aaai25}

\end{document}